\documentclass[11pt]{article}

\usepackage[final]{acl}

\usepackage{times}
\usepackage{latexsym}
\usepackage{booktabs}
\usepackage{multirow}
\usepackage{algorithm}
\usepackage{algpseudocode}

\newcommand{\LINECOMMENT}[1]{\State \textit{// #1}}
\newcommand{\INLINECOMMENT}[1]{\Comment{\textit{#1}}}
\usepackage{amsmath}
\usepackage{amssymb}
\usepackage[T1]{fontenc}
\usepackage{balance}
\usepackage[utf8]{inputenc}

\usepackage{microtype}

\usepackage{inconsolata}

\usepackage{graphicx}

\title{Not All Errors Are Created Equal: ASCoT Addresses Late-Stage Fragility in Efficient LLM Reasoning}

\author{
  \textbf{Dongxu Zhang\textsuperscript{1,}}\thanks{Equal contribution.}, \textbf{Yujun Wu\textsuperscript{2,$\ast$}}, \textbf{Yiding Sun\textsuperscript{1,$\ast$}}, \textbf{Jinnan Yang\textsuperscript{1},} \\  \textbf{Ning Yang\textsuperscript{3}}, \textbf{Jihua Zhu\textsuperscript{1},} 
  \textbf{ Miao Xin\textsuperscript{3}, Baoliang Tian\textsuperscript{4}} \\
  \textsuperscript{1}Xi’an Jiaotong University \quad
  \textsuperscript{2}Peking University \\
  \textsuperscript{3}Institute of Automation, CASIA \quad
  \textsuperscript{4}Tianjin University \\
  \texttt{zhangdongxu@stu.xjtu.edu.cn}
}

\begin{document}
\maketitle
\begin{abstract}

While Chain-of-Thought (CoT) prompting empowers Large Language Models (LLMs), ensuring reasoning reliability remains an open challenge. Contrary to the prevailing cascading failure hypothesis which posits that early errors are most detrimental, we identify a counter-intuitive phenomenon termed \textbf{Late-Stage Fragility}: errors introduced in later reasoning stages are significantly more prone to corrupting final answers. To address this, we introduce ASCoT (Adaptive Self-Correction Chain-of-Thought), a method harmonizing efficiency with robust verification. ASCoT first employs semantic pruning to compress redundant steps, then utilizes an Adaptive Verification Manager (AVM) to prioritize high risk, late-stage steps via a positional impact score, triggering a Multi-Perspective Self-Correction Engine (MSCE) only when necessary. Experiments on GSM8K and MATH-500 demonstrate that ASCoT effectively reallocates computational resources: it reduces token usage by 21\%--30\% for LLaMA-3.1-8B with negligible accuracy drops ($<1.8\%$), achieving a superior trade-off between inference efficiency and reasoning fidelity.

\end{abstract}

\section{Introduction}

\begin{figure}[t]
\centering
\includegraphics[width=0.4\textwidth]{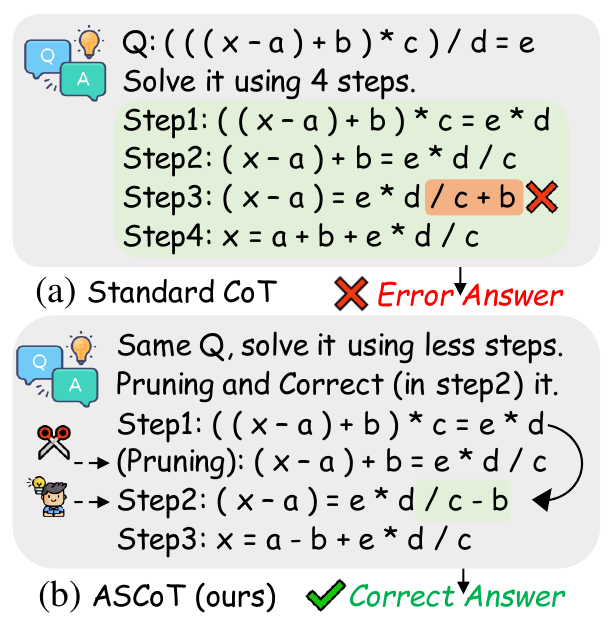} 
\caption{The top part (a) illustrates the standard CoT reasoning process, which often generates redundant outputs and lacks an effective self-correction mechanism when errors occur. The bottom part (b) depicts the ASCoT method, which incorporates a robust mechanism for self-correction, addressing potential reasoning flaws at each stage.}
\label{fig1}
\end{figure}

The advent of CoT has endowed LLMs with remarkable multi-step reasoning capabilities~\cite{kojima2022large}, enabling them to deconstruct complex problems into a sequence of manageable sub-problems~\cite{shinn2023reflexion}. Recent studies, such as the o1 from OpenAI ~\cite{jaech2024openai} and DeepSeekR1 ~\cite{guo2025deepseek} have shown that extended CoT reasoning improves LLM performance. However, the success of CoT and its variants has introduced challenges: prohibitive computational costs from long reasoning chains~\cite{chen2025towards} and the inherent brittleness of the reasoning process itself~\cite{ke2025survey}. A single flaw is often sufficient to invalidate an entire process of thought, raising significant concerns about the trustworthiness of these models~\cite{pan2024llmlingua}.

A widely held belief within the research community, often termed the cascading failure hypothesis, posits that the primary threat to CoT integrity lies in early-stage errors~\cite{tyen2023llms}. The logic is intuitive: an initial mistake in problem comprehension or a miscalculation will inevitably propagate through the logical dependency graph of the reasoning chain, contaminating all subsequent steps. In this work, we directly investigate this foundational assumption, seeking to answer a critical question: Are all errors in a CoT created equal in their impact? To find out this, we conduct a systematic error-injection experiment, which introduces controlled symbolic and numeric errors at varying positions within correct reasoning chains. We then instruct the LLM to resume the reasoning process from the point at which the error occurred. Our findings are striking and counter-intuitive. We discover that errors introduced in later stages are more likely to lead to an incorrect final answer. We term this phenomenon Late-Stage Fragility. 
An early error often appears to trigger the model's latent self-correction mechanisms, allowing it to recover and still reach the correct solution. 

As the model progresses steadily, its reasoning process gains a form of semantic commitment, reducing relevant information entropy. At the later stages, the model is unable to evaluate the validity of the final calculations.
This empirical discovery suggests that a significant portion of efforts to secure CoT reasoning may be misdirected. As illustrated in Figure~\ref{fig1}, a standard CoT process can be both inefficient due to redundancy and unreliable due to its inability to correct these late-stage errors. Building on a rethinking of how we evaluate and how we correct LLM reasoning, we introduce the ASCoT method. In contrast to standard methods, ASCoT is designed to address both challenges simultaneously, it first employs a semantic pruning stage to enhance efficiency, followed by a targeted, risk-aware correction mechanism that focuses computational resources precisely on the most vulnerable steps to ensure a correct final answer.

Our contributions are summarized as follows:
\begin{itemize}
    \item To the best of our knowledge, this is the first to identify and quantify the Late-Stage Fragility phenomenon in CoT reasoning, demonstrating that errors in later steps are significantly more detrimental than those in earlier steps.
    \item We introduce ASCoT, a novel method specifically designed to counteract this fragility by precisely identifying and robustly correcting the high-impact, late-stage errors.
    \item Our extensive experiments validate the effectiveness of ASCoT. We demonstrate that ASCoT achieves outstanding accuracy on the GSM8k and MATH benchmarks.
\end{itemize}

\section{Related Work}

\subsection{LLM Reasoning \& Efficient Thinking}

CoT prompting significantly advances LLM reasoning by decomposing complex problems into intermediate steps~\cite{liu2024can,team2024qwen2}. To mitigate the associated computational costs, recent research explores diverse efficiency-focused strategies, including prompt-based constraints ($e.g.,$ token budgets)~\cite{xu2025chain,han2024token}, inference-time dynamic early-exit mechanisms~\cite{ding2025dynamic}, and CoT distillation~\cite{wadhwa2024investigating,zhang2026chain}.

\subsection{CoT Compression and Pruning}

The inherent verbosity of CoT incurs significant computational overhead~\cite{wingate2022prompt} and can induce the overthinking phenomenon, where excessive reasoning introduces errors~\cite{kumar2025overthink}. To mitigate this, recent research focuses on compression strategies~\cite{xuefei2023skeleton}. Approaches vary from token-level pruning based on semantic importance~\cite{zhu2025breaking,yang2025token} to structural methods like Prune-on-Logic~\cite{liu2025efficient}, which filters steps via logic graphs. Notably, even heuristic truncation~\cite{wang2024svd} has demonstrated efficacy, suggesting that exhaustive reasoning paths are often redundant.

\subsection{Self-correction Mechanism}

Self-correction mechanisms enhance LLM reliability by iterating through generation, feedback, and refinement phases~\cite{kamoi2024can}. Current approaches bifurcate based on the feedback source. Intrinsic self-correction relies on the model's internal critique capabilities without external tools~\cite{yang2025token,zhu2025breaking}. However, these methods often struggle to identify errors due to the scarcity of correction examples in training data~\cite{liu2024intrinsic}. In contrast, extrinsic self-correction leverages external signals, such as verifier models~\cite{kim2025search} or consistency checks, to rigorously assess step-wise correctness.

\section{Preliminary}
This section establishes the foundational concepts that motivate our work, offering a review of some effective methods for identifying token importance.

\textbf{Efficient CoT in LLM Reasoning}\quad A critical question for CoT efficiency is whether every token in different position contributes equally to the final answer~\cite{xia2025tokenskip}. Several early approaches~\cite{hou2022token} measure importance based on the self-information or surprisal from a unidirectional language model:
\begin{equation}
    I_1(x_i) = -\log P(x_i | x_{< i}; \theta_{\mathcal{M}_L}) ,\quad
    \label{eq1}
\end{equation}
where $x_i$ is a given token, $\theta_{\mathcal{M}_L}$ denotes the LLM used to compute the confidence and $P(\cdot\mid \cdot;\theta_{\mathcal{M}_L})$ is the conditional probability assigned by the causal LLMs. This metric suffers from known limitations, such as positional dependency, where tokens at the end of sentences often receive lower importance scores due to the nature of perplexity in LLMs.

To address this, more advanced methods like LLMLingua-2~\cite{pan2024llmlingua} utilize a bidirectional language model~\cite{devlin2019bert} ($e.g.,$ BERT-style) to reframe the task as token classification. In this paradigm, the importance of a token $x_i$ is defined as the probability that a trained classifier assigns it an essential label ($y_i=1$), conditioned on the entire sequence $x$:
\begin{equation}
I_2(x_i) = P(y_i=1 | x; \theta_{\mathcal{M}{cls}}),
\label{eq2}
\end{equation}
where $\theta_{\mathcal{M}{cls}}$ represents the parameters of the trained bidirectional classifier. Such methods confirm that tokens within a CoT sequence are variable in importance. For example, tokens comprising mathematical equations or key entities~\cite{ma2025step} are identified as critical, whereas semantic connectors such as ``so'' and ``since'' contribute less to the final outcome of the reasoning ~\cite{hou2504thinkprune}. This highlights that CoT efficiency can be improved by pruning these redundant tokens.


\textbf{Error Propagation in Reasoning Process}\quad Beyond efficiency, the primary challenge for CoT is reliability. A single error can invalidate an entire reasoning process~\cite{guo2017critical}. The conventional understanding of this problem is rooted in the cascading failure hypothesis, which posits that errors are most damaging when they occur early in the reasoning chain~\cite{jiang2025cascadia}. According to this view, an early mistake propagates through the logical dependency graph, contaminating all subsequent steps.
In this study, through systematic error-injection experiments, we observe that errors introduced in the later stages of a reasoning chain are significantly more detrimental to the final outcome than identical errors introduced at the beginning.

\begin{figure*}[t]
\centering
\includegraphics[width=0.95\textwidth]{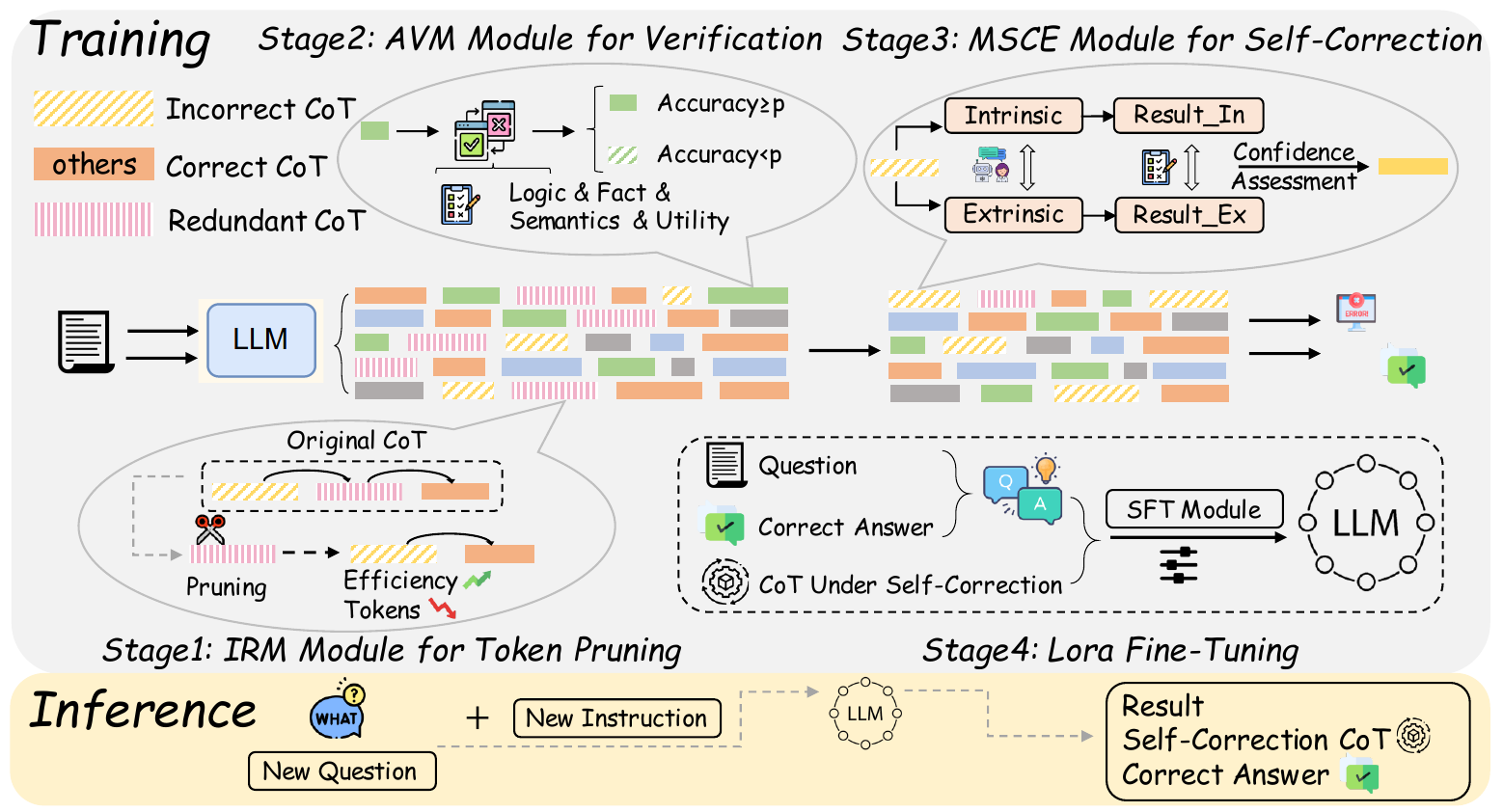} 
\caption{Illustration of ASCoT Pipeline. ASCoT generates the $CoT_{initial}$ from the target LLM, which is then compressed by IRM module to a predefined ratio $\gamma$. The compressed CoT is verified by AVM module, and if the confidence score exceeds a threshold $\tau$, the problematic steps are sent to MSCE module for error correction. The corrected $CoT_{final}$ is then used for fine-tuning. Finally, ASCoT enables reasoning on new problems with the specified compression ratio $\gamma$.}
\label{fig2}
\end{figure*}

\section{Method}
\label{others}

In this section, we detail the architecture of our method. 
ASCoT generates the $CoT_{initial}$ based on the target LLM. The $CoT_{initial}$ is then processed through the Intelligent Routing Mechanism (IRM) module, reducing the output to a fixed ratio $\gamma$. The remaining CoT is passed through the Adaptive Verification Manager (AVM) module, which performs a rigorous verification process. If the model's risk score for any CoT step exceeds a predefined threshold $\tau$, indicating a potential error, the corresponding CoT step is forwarded to the Multi-Perspective Self-Correction Engine (MSCE) module. The MSCE module applies internal and external validation techniques to correct errors, resulting in the corrected $CoT_{final}$. Ultimately, the refined, non-redundant $CoT_{final}$ is used for fine-tuning. We illustrate the training and inference process in Figure~\ref{fig2}, respectively. We also provide Algorithm~\ref{alg:ASCoT} for ASCoT.

\subsection{Intelligent Routing Mechanism}
The IRM serves as the initial processing stage, responsible for preliminary CoT pruning and compression. The core idea is inspired by methods TokenSkip~\cite{xia2025tokenskip}, which recognize that not all tokens in a CoT are equally important for the final reasoning outcome.
The mechanism operates by first generating a complete CoT path for a given problem. Then it employs a trained token importance model to assign a semantic importance score to each token in the chain. Based on a predefined compression ratio $\gamma$, tokens with scores below the corresponding threshold are pruned, resulting in a compressed CoT, namely the final answer.

\subsection{Adaptive Verification Manager}
AVM dynamically identifies which steps are most likely to be incorrect and then computes a Risk Score ${R}(t_k)$ for each step $t_k$ by integrating two key assessments: the step's confidence assessment $Q(t_k)$ and its positional impact $I(k)$.

\textbf{Confidence Assessment Score}\quad
To provide a holistic measure of the single reasoning step's quality, we first define a comprehensive quality score $Q(t_k)$, which is a composite metric that evaluates step $t_k$ across four dimensions: Logical Validity $V(t_k)$, Factual Support $G(t_k)$, Semantic Clarity $C(t_k)$ and Process Utility $U(t_k)$. 

Logical validity $V(t_k)$ evaluates the internal logical consistency of step $t_k$. The score is a binary value that indicates if the step $t_k$ is derivable from its context $(P,S_{<k})$, determined by a verification query to the LLM $\mathcal{M}_L$:
\begin{equation}
V(t_k) = \mathbb{I}(o_k^v \in y),
\end{equation}
where $o_k^v$ is the model's output of step $t_k$, $y$ is the set of valid responses and $\mathbb{I}(\cdot)$ is the indicator function.

Factual support $G(t_k)$ measures the correctness of all $n$ verifiable assertions within the step. For the mathematical reasoning benchmarks used in this work, verifiable assertions are arithmetic calculations. The function is defined as:
\begin{equation}
G(t_k) = \prod_{i=1}^{n} T(a_i),
\end{equation}
where $a_i$ is the $i$-th arithmetic operation in step $t_k$, and $T(a_i)\rightarrow\{0,1\}$ is a verification function that executes the operation using a computational engine ($e.g.,$ a Python interpreter) to confirm its accuracy.

Semantic clarity $C(t_k)$ quantifies the coherence of step $t_k$. To ensure the score is well-calibrated and bounded within a $[0, 1]$ range, we normalize the average conditional log-probability of its tokens $\{y_1,...,y_m\}$ using a sigmoid function:
\begin{equation}
C(t_k) = \sigma\left(\frac{1}{m} \sum_{i=1}^{m} \log P(y_i | c_{<i})\right),
\end{equation}
where $\sigma(\cdot)$ is the sigmoid function, $y_i$ is the $i$-th token of step $t_k$, and $c_{<i}$ is the preceding context. 

Process utility $U(t_k)$ measures the contribution of step $t_k$ to the final answer $\mathcal{A}^*$. It is defined as the information gain:
\begin{equation}
U(t_k) = \log P(\mathcal{A}^* | S_{\le k}) - \log P(\mathcal{A}^* | S_{<k}), 
\end{equation}
where $\mathcal{A}^*$ is the ground‐truth answer, $S_{\le k}$ is the full context including step $t_k$ and $S_{< k}$is the context excluding step $t_k$. $U(t_k)$ relies on the ground-truth answer and is used during the training phase. At inference time, this term is omitted.



\begin{figure}[t]
\centering
\includegraphics[width=0.4\textwidth]{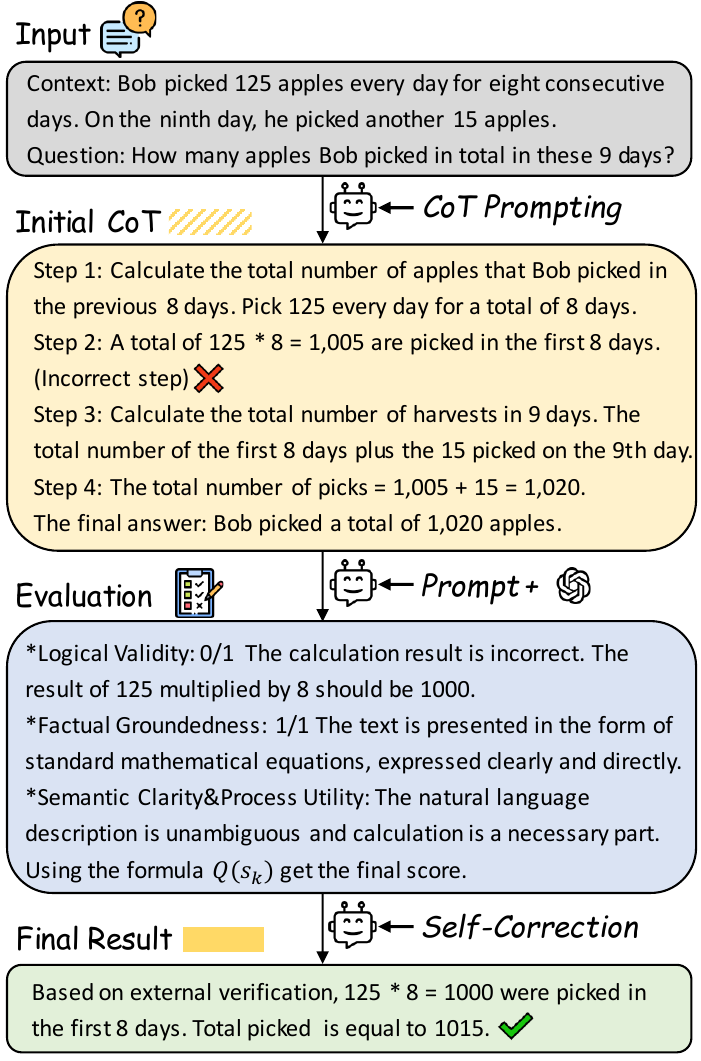} 
\caption{The details of verification and self-correction. AVM and MSCE working together to correct a faulty step $t_k$. AVM computes $R(t_k)$. MSCE then applies intrinsic and extrinsic correction to produce the final corrected $CoT_{final}$.}
\label{fig6}
\end{figure}

Based on these dimensions, we design the single-step comprehensive quality scoring function $Q(t_k)$ with a veto mechanism for foundational criteria:
\begin{equation}
  Q(t_k)
  = V(t_k) \cdot G(t_k) \cdot
    \left(
      w_c \cdot C(t_k)
      + w_u \cdot U(t_k)
    \right),
  \label{eq4}
\end{equation}
where $w_c$ and $w_u$ are hyperparameters that balance the contribution of clarity and utility. Figure~\ref{fig6} illustrates this evaluation process, where an initial CoT with a calculation error is analyzed.



\textbf{Positional Impact Score}\quad
To quantitatively model the Late-Stage Fragility phenomenon observed in our error-injection experiments (Section 4.3), we introduce the Position Impact Function $I(k)$. This function is an empirical model fitted to our experimental data, designed to capture the escalating impact of errors at later reasoning stages. We normalize this position function as:
\begin{equation}
  I(k) = w_a \cdot e^{\alpha (k/K)},
  \label{eq5}
\end{equation}
where $k$ is the step's index, $K$ is the total number of steps, $w_a$ is a base impact coefficient, and $\alpha$ is a propagation rate. These parameters are determined by fitting the function to our error-injection data. This procedure involves 
injecting single errors at different steps $k$ in a set of baseline correct reasoning chains and measuring the probability of final answer failure. The function $I(k)$ is then fitted to these empirical data points. Details of the analysis are available in Appendix~\ref{appen:error}.

Finally, AVM combines these factors into a single Risk Score:
\begin{equation}
  {R}(t_k) = I(k)\,\times\bigl(1 - Q(t_k)\bigr).
  \label{eq6}
\end{equation}

This score is then compared against a tunable system threshold $\tau$. If $R(t_k)>\tau$, the step is flagged as high risk and passed to the MSCE.

\subsection{Multi-Perspective Self-Correction Engine}
Once a step $t_k$ is flagged by the AVM, the MSCE is activated to perform a robust and efficient correction. To counteract a model struggles to correct its own mistakes when using a single line of reasoning, the MSCE employs a dual-path correction strategy: Intrinsic Correction, the model is presented with the context up to step $s_{(k-1)}$ and its own flawed step $t_k$, and is explicitly prompted to review and amend it. Extrinsic Correction, the model is given the context up to $s_{(k-1)}$ but without the flawed step $t_k$. 
Upon generating two candidate corrections, the MSCE employs a selection mechanism to choose the superior one. We re-apply the Confidence Assessment Score function $Q(\cdot)$ (from Eq.~\ref{eq4}, with $U=0$) to both candidates. The candidate with the higher quality score is then integrated into the reasoning chain. 

\begin{table*}[t]
\centering
\caption{Experimental results of ASCoT on LLaMA-3.1-8B-Instruct. We report Accuracy, average CoT token count (Tokens) and actual compression ratio (ActRatio) for comparison. The compression ratio $\gamma$ is set to $0.5-1.0$.}
\begin{tabular}{l c 
                *{3}{c} 
                *{3}{c}}
\toprule
\multirow{2}{*}{\textbf{Methods}} 
  & \multirow{2}{*}{\textbf{Ratio}} 
  & \multicolumn{3}{c}{\textbf{GSM8K}} 
  & \multicolumn{3}{c}{\textbf{MATH-500}} \\
\cmidrule(lr){3-5} \cmidrule(lr){6-8}
  & 
  & \textbf{Accuracy\,$\uparrow$} 
  & \textbf{Tokens\,$\downarrow$} 
  & \textbf{ActRatio} 
  & \textbf{Accuracy\,$\uparrow$} 
  & \textbf{Tokens\,$\downarrow$} 
  & \textbf{ActRatio} \\
\midrule
Original 
& 1.0 & 86.2\scriptsize(\textcolor{red}{0.0$\downarrow$})  & 213.17 & 1.00 & 48.6\scriptsize(\textcolor{red}{0.0$\downarrow$}) & 502.60 &  1.00 \\
\midrule
\multirow{3}{*}{Prompt}
& 0.9 & 84.1\scriptsize(\textcolor{red}{2.1$\downarrow$}) & 226.37 & 1.06 & 48.6\scriptsize(\textcolor{red}{0.0$\downarrow$}) & 468.04 & 0.93 \\
& 0.7 & 84.9\scriptsize(\textcolor{red}{1.3$\downarrow$}) & 209.39 & 0.98 & 48.4\scriptsize(\textcolor{red}{0.4$\downarrow$}) & 472.13 & 0.94 \\
& 0.5 & 83.7\scriptsize(\textcolor{red}{2.5$\downarrow$}) & 188.82 & 0.89 & 47.8\scriptsize(\textcolor{red}{0.4$\downarrow$}) & 471.11 & 0.94 \\
\midrule
\multirow{3}{*}{Truncation} 
& 0.9 & 70.2\scriptsize(\textcolor{red}{26.0$\downarrow$}) & 202.06 & 0.95 & 47.8\scriptsize(\textcolor{red}{0.8$\downarrow$}) & 440.33 & 0.88 \\
& 0.7 & 25.9\scriptsize(\textcolor{red}{60.3$\downarrow$}) & 149.97 & 0.70 & 45.0\scriptsize(\textcolor{red}{3.6$\downarrow$})& 386.89 & 0.77 \\
& 0.5 & 7.0\scriptsize(\textcolor{red}{79.2$\downarrow$}) & 103.69 & 0.49 & 27.4\scriptsize(\textcolor{red}{21.2$\downarrow$})& 283.70 & 0.56 \\
\midrule
\multirow{6}{*}{ASCoT} 
& 1.0 & 86.9\scriptsize(\textcolor{green}{0.7$\uparrow$}) & 214.24 & 1.00 & 48.8\scriptsize(\textcolor{green}{0.2$\uparrow$}) & 478.80 & 0.95 \\
& 0.9 & 86.1\scriptsize(\textcolor{red}{0.1$\downarrow$}) & 197.65 & 0.92 & 47.6\scriptsize(\textcolor{red}{1.0$\downarrow$}) & 427.15 & 0.84 \\
& 0.8 & 85.1\scriptsize(\textcolor{red}{1.1$\downarrow$}) & 169.96 & 0.79 & 47.2\scriptsize(\textcolor{red}{1.4$\downarrow$}) & 376.72 & 0.74 \\
& 0.7 & 83.7\scriptsize(\textcolor{red}{2.5$\downarrow$}) & 151.32 & 0.71 & 46.8\scriptsize(\textcolor{red}{1.8$\downarrow$}) & 334.26 & 0.67 \\
& 0.6 & 81.6\scriptsize(\textcolor{red}{4.6$\downarrow$}) & 128.38 & 0.60 & 41.8\scriptsize(\textcolor{red}{6.8$\downarrow$}) & 309.69 & 0.61 \\
& 0.5 & 79.5\scriptsize(\textcolor{red}{6.7$\downarrow$}) & 118.57 & 0.55 & 40.1\scriptsize(\textcolor{red}{8.5$\downarrow$}) & 288.94 & 0.57 \\
\bottomrule
\end{tabular}
\label{table1}
\end{table*}

\section{Experiment}
\subsection{Experimental Setup}
\textbf{Models and Datasets}\quad We primarily evaluate our method using the LLaMA-3.1-8B-Instruct~\cite{dubey2024llama} and Qwen2.5-Instruct series (3B, 7B, 14B)~\cite{team2024qwen2}models. Our evaluation leverages two notably standard math reasoning benchmarks: GSM8K~\cite{cobbe2021training}, which consists of 8.5K grade school math problems requiring 2-8 reasoning steps, and MATH~\cite{hendrycks2021measuring}, a more challenging dataset of 12.5K competition-level math problems. Given the significant computational cost associated with the MATH dataset, we follow recent research practices by evaluating our method on a carefully selected subset, MATH-500~\cite{lightman2023let}, which includes a representative sample of 500 problems.

\textbf{Implementation Details}\quad The ASCoT is implemented in PyTorch 1.13.0 and all experiments are conducted on a server with four NVIDIA GeForce RTX 3090 GPUs (24GB). For the IRM module, we use LLMLingua-2~\cite{pan2024llmlingua} as the token importance metric to generate compressed CoT data for fine-tuning. To optimize the fine-tuning process, we adopt LoRA~\cite{hu2022lora} for all fine-tuning procedures, setting the rank $r=8$ and scaling parameter $\alpha=16$ to ensure efficient training. Additional implementation details and configurations can be found in the Appendix~\ref{appen::Exset}.

\textbf{Evaluation Metrics}\quad We evaluate the performance of ASCoT using three primary metrics: Accuracy (\%), which represents the percentage of problems for which the final answer is correct. Tokens, the average number of tokens generated in the reasoning chain, measuring computational cost and efficiency during the reasoning process. ActRatio, the actual compression ratio of ASCoT, reflecting the real efficiency of the token compression.

\begin{figure}[t]
\centering
\includegraphics[width=0.4\textwidth]{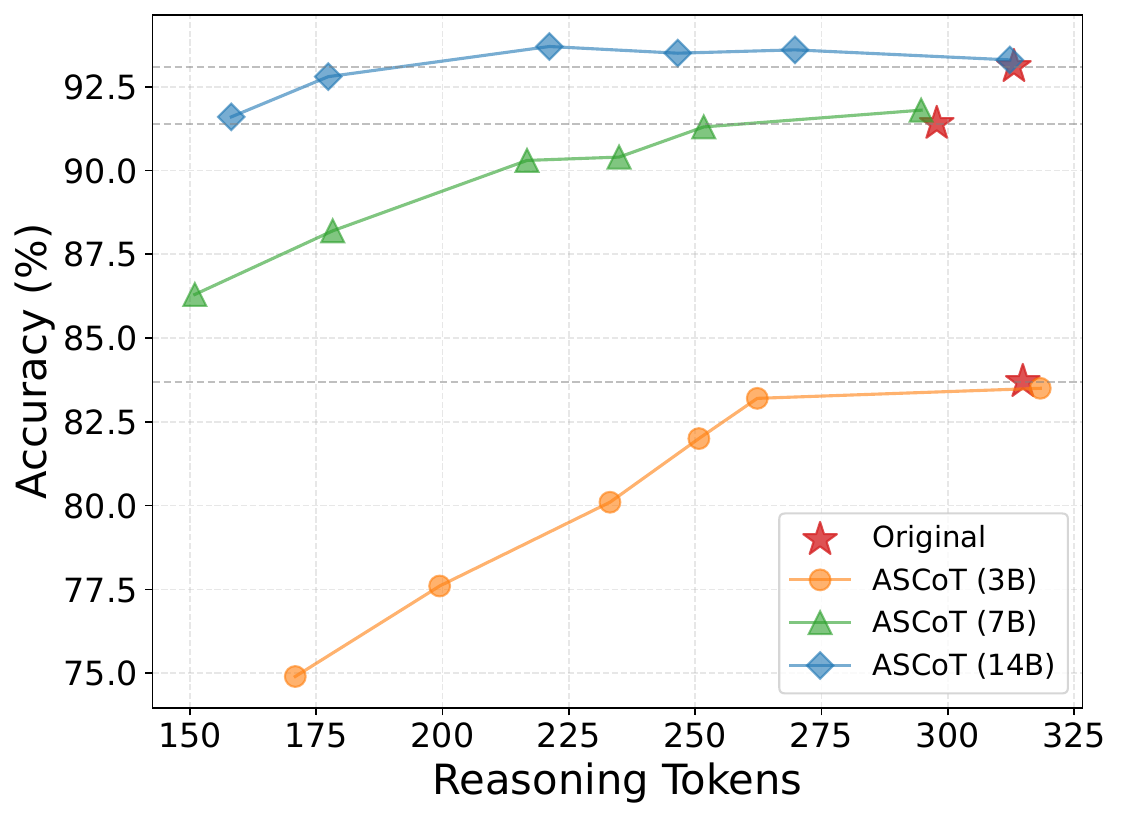} 
\caption{Performance of ASCoT on the Qwen2.5-Instruct Series. The results for the 3B, 7B, and 14B models are shown under various compression ratios and compared against the original, uncompressed baseline.}
\label{fig5}
\end{figure}

\subsection{Main Results}

We benchmark ASCoT against two standard length-control baselines: instruction-based Prompting and hard Truncation. The comparative results on LLaMA-3.1-8B are presented in Table \ref{table1}.

\textbf{Comparison with Baselines}\quad
Existing methods struggle to balance compression controllability with reasoning fidelity. Prompting fails to strictly adhere to token budgets. Truncation also incurs catastrophic performance degradation when it enforces strict length constraints.

In contrast, ASCoT demonstrates precise controllability and robust performance. It aligns closely with target ratios (e.g., ActRatio 0.79 at a 0.8 target) while preserving reasoning accuracy. Notably, at a full budget ($\gamma=1.0$), ASCoT marginally outperforms the original baseline (+0.7\% on GSM8K), suggesting that our verification mechanism effectively filters noise. Even under aggressive compression ($\gamma=0.5$), ASCoT maintains a respectable 79.5\% accuracy on GSM8K, significantly outperforming truncation based methods.

\textbf{Scalability Analysis}\quad
Table~\ref{table2} details the performance across the Qwen2.5-Instruct series (3B, 7B, 14B) to evaluate scalability. ASCoT achieves substantial token reduction with minimal accuracy loss across all scales. For the 3B model, a 0.9 compression ratio reduces token usage by approximately 16\% with only a 0.5 point accuracy drop. 
This resilience becomes more pronounced as model size increases. The 14B model exhibits exceptional robustness. Even when halving the token budget, it retains 91.6\% accuracy (a mere 1.5\% drop). This trend suggests a positive correlation between model scale and reasoning redundancy, larger models tend to generate more steps that ASCoT effectively identifies and prunes without compromising the final answer, demonstrating its potential to significantly lower inference costs for large-scale deployments without sacrificing reliability.

\begin{figure*}[t]
\centering
\includegraphics[width=0.65\textwidth]{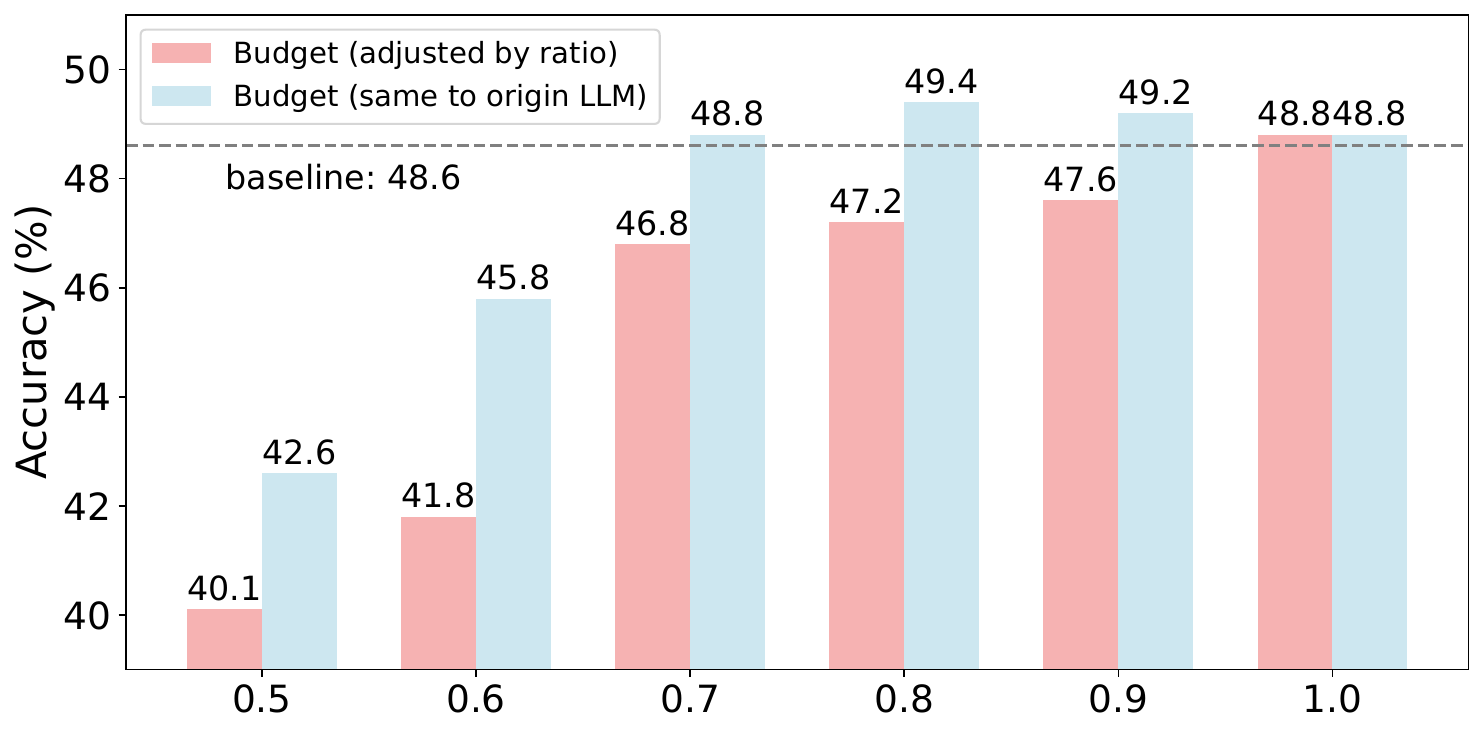} 
\caption{Performance comparison of ASCoT with varying maximum length constraints. We adjust the maximum length budget when evaluating our method on MATH-500.}
\label{fig3}
\end{figure*}

Figure~\ref{fig5} illustrates ASCoT's robust scalability across the Qwen2.5-Instruct series. While the 3B model incurs a negligible 0.5-point accuracy drop at 17\% compression, this resilience amplifies with model size. The 7B variant maintains over 86\% accuracy even with a 50\% token reduction. Most notably, the 14B model demonstrates exceptional efficiency retention: halving the token budget results in a mere 1.5\% accuracy decline. This trend indicates a positive correlation between model capacity and reasoning redundancy, suggesting that ASCoT effectively exploits the richer semantic structures inherent in larger models to maximize pruning without information loss. This confirm ASCoT's capability to balance computational efficiency and reasoning fidelity across varying model scales.

\subsection{Error Propagation of Controlled Injection}

To interrogate the cascading failure hypothesis, we conducted a controlled error-injection study on GSM8K using Qwen2.5-7B. We injected two classes of perturbations mimicking naturally occurring mistakes: Numeric (value deviation) and Symbolic (operator inversion). 

Results in Table \ref{table3} reveal a Late-Stage Fragility. Contrary to the expectation that early errors propagate most severely, we observe that impact escalates with position. For numeric errors, a miscalculation at the final step (4/4) causes a catastrophic 51.69\% accuracy drop, significantly outpacing early-stage errors (14.64\% at Step 2). Symbolic errors amplify this trend, with late-stage injections (Step 3/3) causing the most profound degradation (56.78\%). These findings suggest that models become increasingly committed to their reasoning trajectory, losing the capacity to self-correct as they approach the solution. Distribution details of the inference steps are provided in Table \ref{table5}.

\begin{table}[t]
\centering
\setlength{\tabcolsep}{3.5pt} 
\renewcommand{\arraystretch}{0.9} 

\caption{Experimental results of the Qwen2.5-Instruct series evaluated on GSM8K. We report Accuracy (\%), Tokens and ActRatio.}

\resizebox{\columnwidth}{!}{%
    \begin{tabular}{l c c c c c}
    \toprule
    \textbf{Scale} & \textbf{Methods} & \textbf{Ratio} & \textbf{Accuracy} & \textbf{Tokens} & \textbf{ActRatio} \\
    \midrule
    \multirow{7}{*}{3B}
      & Original & -- & 83.7\scriptsize(\textcolor{red}{0.0$\downarrow$}) & 314.87 & -- \\ 
      \cmidrule(lr){2-6}
      & \multirow{6}{*}{ASCoT}
        & 1.0 & 83.5\scriptsize(\textcolor{red}{0.2$\downarrow$}) & 318.32 & 1.01 \\ 
      & & 0.9 & 83.2\scriptsize(\textcolor{red}{0.5$\downarrow$}) & 262.31 & 0.83 \\ 
      & & 0.8 & 82.0\scriptsize(\textcolor{red}{1.7$\downarrow$}) & 250.75 & 0.79 \\ 
      & & 0.7 & 80.1\scriptsize(\textcolor{red}{3.6$\downarrow$}) & 233.15 & 0.74 \\ 
      & & 0.6 & 77.6\scriptsize(\textcolor{red}{6.1$\downarrow$}) & 199.44 & 0.63 \\ 
      & & 0.5 & 74.9\scriptsize(\textcolor{red}{8.8$\downarrow$}) & 170.87 & 0.54 \\
    \midrule
    \multirow{7}{*}{7B}
      & Original & -- & 91.4\scriptsize(\textcolor{red}{0.0$\downarrow$}) & 297.83 & -- \\ 
      \cmidrule(lr){2-6}
      & \multirow{6}{*}{ASCoT}
        & 1.0 & 91.8\scriptsize(\textcolor{green}{0.4$\uparrow$}) & 294.74 & 0.99 \\ 
      & & 0.9 & 91.3\scriptsize(\textcolor{red}{0.1$\downarrow$}) & 251.71 & 0.85 \\ 
      & & 0.8 & 90.4\scriptsize(\textcolor{red}{1.0$\downarrow$}) & 234.95 & 0.79 \\ 
      & & 0.7 & 90.3\scriptsize(\textcolor{red}{1.1$\downarrow$}) & 216.73 & 0.73 \\ 
      & & 0.6 & 88.2\scriptsize(\textcolor{red}{3.2$\downarrow$}) & 178.28 & 0.60 \\ 
      & & 0.5 & 86.3\scriptsize(\textcolor{red}{5.1$\downarrow$}) & 150.98 & 0.51 \\
    \midrule
    \multirow{7}{*}{14B}
      & Original & -- & 93.1\scriptsize(\textcolor{red}{0.0$\downarrow$}) & 313.11 & -- \\ 
      \cmidrule(lr){2-6}
      & \multirow{6}{*}{ASCoT}
        & 1.0 & 93.3\scriptsize(\textcolor{green}{0.2$\uparrow$}) & 312.30 & 0.99 \\ 
      & & 0.9 & 93.6\scriptsize(\textcolor{green}{0.5$\uparrow$}) & 269.76 & 0.86 \\ 
      & & 0.8 & 93.5\scriptsize(\textcolor{green}{0.4$\uparrow$}) & 246.57 & 0.79 \\ 
      & & 0.7 & 93.7\scriptsize(\textcolor{green}{0.6$\uparrow$}) & 221.17 & 0.71 \\ 
      & & 0.6 & 92.8\scriptsize(\textcolor{red}{0.3$\downarrow$}) & 177.40 & 0.57 \\ 
      & & 0.5 & 91.6\scriptsize(\textcolor{red}{1.5$\downarrow$}) & 158.21 & 0.51 \\
    \bottomrule
    \end{tabular}%
}
\label{table2}
\end{table}

\subsection{Analysis}
\textbf{Length Budget Analysis}\quad 
To further investigate ASCoT's efficiency, we analyze its performance under a fixed token budget. As illustrated in Figure~\ref{fig3}, we assess our approach under two distinct length-budget settings. In the first, the maximum generation length is scaled by the compression ratio (ranging from 0.5 to 1.0). In the second, we retain the original LLM budget ($e.g.,$ 512 tokens in GSM8K and 1024 tokens in MATH) regardless of compression. The results reveal that under the unchanged budget, our method consistently surpasses the baseline: accuracy rises from 42.6\% to a peak of 49.4\%. Notably, at compression ratios of 0.8 and 0.9, we observe absolute accuracy gains of 0.8 and 0.6 percentage points over the baseline in a smaller tokens, underscoring the potential of compressed CoT reasoning within the same budget.

\textbf{The Late-Stage Fragility of CoT}\quad 
Our findings refine the traditional cascading failure model by uncovering Late-Stage Fragility phenomenon. While early-stage errors often trigger latent self-correction mechanisms due to high contextual entropy, identical errors introduced in the final steps are significantly more prone to persist uncorrected (qualitative illustration provided in Appendix~\ref{sec:app_fragility}). This suggests that the model forms a rigid semantic commitment as reasoning progresses. Consequently, while cascading failures undeniably exist, Late-Stage Fragility constitutes a distinct, critical failure mode where the model lacks the flexibility to revisit and rectify terminal errors.

\textbf{Case Study}\quad 
To provide a clear and detailed qualitative understanding of our method, we present a case study. Figure~\ref{fig6} illustrates a qualitative analysis where a standard model correctly structures the reasoning but commits a critical arithmetic error by calculating $125 \times 8$ as $1005$. This flaw propagates to an incorrect conclusion. ASCoT effectively addresses this issue as the AVM module flags the step due to low Logical Validity while acknowledging its structural soundness. Crucially, this granular diagnosis prevents the unnecessary discarding of the entire reasoning chain. Instead, the MSCE executes a targeted dual-path repair to synthesize a corrected candidate, rectifying the calculation to $1000$, thereby enabling the model to recover and deduce the correct total of $1015$.

\begin{table}[t]
\centering
\renewcommand{\arraystretch}{0.9}
\caption{Accuracy for Different Inference Steps in the GSM8K Dataset, using Qwen2.5-Instruct 7B.}
\begin{tabular}{c c c c }
\toprule
\textbf{Steps} & \textbf{Correct} & \textbf{Total} & \textbf{Accuracy} \\ 
\midrule
0 & 615 & 660 & 93.18 \\ 
1 & 1 & 1 & 100.00 \\ 
2 & 777 & 786 & 98.85 \\ 
3 & 2044 & 2125 & 96.19 \\ 
4 & 1936 & 2022 & 95.75 \\ 
5 & 1110 & 1180 & 94.07 \\ 
6 & 440 & 484 & 90.91 \\ 
7 & 117 & 137 & 85.40 \\ 
8 & 39 & 58 & 67.24 \\
9--11 & 17 & 20 & 85.00 \\
\bottomrule
\end{tabular}
\label{table5}
\end{table}

\subsection{Ablation Study}

To rigorously evaluate the contribution of each component within ASCoT, we conducted a multi-level ablation study on the GSM8K dataset. We first validated the necessity of the core modules: IRM and AVM. Subsequently, we performed a fine-grained analysis of the MSCE module to disentangle the efficacy of intrinsic versus extrinsic correction strategies. Our results detailed in Appendix~\ref{sec:app_ablation}, confirm that the synergistic integration of position-aware verification and dual-path correction is essential for achieving state-of-the-art performance while minimizing token usage.

\begin{table}[t]
\centering
\setlength{\tabcolsep}{4pt} 
\renewcommand{\arraystretch}{0.85} 

\caption{Error-injection results on GSM8K using Qwen2.5-Instruct 7B. ``All/Err'' denotes the total steps vs. injected step. ``Drop'' denotes the accuracy drop.}

\resizebox{\columnwidth}{!}{%
    \begin{tabular}{c c c c c}
    \toprule
    \textbf{All/Err} & \textbf{Type} & \textbf{Ori Acc} & \textbf{Fin Acc} & \textbf{Drop} \\
    \midrule
    4/2 & \multirow{3}{*}{numeric}  & 95.75 & 81.11 & 14.64 \\
    4/3 &                           & 95.75 & 66.19 & 29.56 \\
    4/4 &                           & 95.75 & 44.06 & 51.69 \\
    \midrule
    2/2 & \multirow{3}{*}{symbolic} & 98.85 & 54.88 & 43.97 \\
    3/2 &                           & 96.19 & 78.28 & 17.91 \\
    3/3 &                           & 96.19 & 39.41 & 56.78 \\
    \bottomrule
    \end{tabular}%
}
\label{table3}
\end{table}

\section{Conclusion}

In this work, we critically re-examine the dynamics of error propagation in CoT reasoning, identifying a counter-intuitive phenomenon termed Late-Stage Fragility. Our systematic analysis reveals that contrary to the prevailing cascading failure assumption, errors introduced in the terminal stages of reasoning impose a disproportionately higher penalty on the final outcome. 
To address this vulnerability, we introduce ASCoT (Adaptive Self-Correction Chain-of-Thought). By integrating an Intelligent Routing Mechanism with a position-aware Adaptive Verification Manager, ASCoT strategically reallocates computational resources to verify high risk, late-stage steps. Extensive experiments on GSM8K and MATH benchmarks demonstrate that ASCoT achieves a superior trade-off between reasoning fidelity and computational efficiency, outperforming strong baselines including Self-Consistency. Ultimately, our findings advocate for a paradigm shift from uniform verification to adaptive, context-sensitive strategies, establishing a new foundation for reliable and resource-efficient LLM reasoning.

\section*{Limitations}
While ASCoT effectively mitigates Late-Stage Fragility, its performance naturally correlates with the base model's verification capabilities. Additionally, our current study primarily targets mathematical reasoning focusing on numeric and symbolic errors. Extending the framework to broader domains ($e.g.,$ code generation) and diverse modalities remains a promising avenue for future exploration.




\bibliography{custom}

@article{kojima2022large,
  title={Large language models are zero-shot reasoners},
  author={Kojima, Takeshi and Gu, Shixiang Shane and Reid, Machel and Matsuo, Yutaka and Iwasawa, Yusuke},
  journal={Advances in neural information processing systems},
  volume={35},
  pages={22199--22213},
  year={2022}
}

@article{shinn2023reflexion,
  title={Reflexion: Language agents with verbal reinforcement learning},
  author={Shinn, Noah and Cassano, Federico and Gopinath, Ashwin and Narasimhan, Karthik and Yao, Shunyu},
  journal={Advances in Neural Information Processing Systems},
  volume={36},
  pages={8634--8652},
  year={2023}
}

@article{jaech2024openai,
  title={Openai o1 system card},
  author={Jaech, Aaron and Kalai, Adam and Lerer, Adam and Richardson, Adam and El-Kishky, Ahmed and Low, Aiden and Helyar, Alec and Madry, Aleksander and Beutel, Alex and Carney, Alex and others},
  journal={arXiv preprint arXiv:2412.16720},
  year={2024}
}

@article{guo2025deepseek,
  title={Deepseek-r1: Incentivizing reasoning capability in llms via reinforcement learning},
  author={Guo, Daya and Yang, Dejian and Zhang, Haowei and Song, Junxiao and Zhang, Ruoyu and Xu, Runxin and Zhu, Qihao and Ma, Shirong and Wang, Peiyi and Bi, Xiao and others},
  journal={arXiv preprint arXiv:2501.12948},
  year={2025}
}

@article{chen2025towards,
  title={Towards reasoning era: A survey of long chain-of-thought for reasoning large language models},
  author={Chen, Qiguang and Qin, Libo and Liu, Jinhao and Peng, Dengyun and Guan, Jiannan and Wang, Peng and Hu, Mengkang and Zhou, Yuhang and Gao, Te and Che, Wanxiang},
  journal={arXiv preprint arXiv:2503.09567},
  year={2025}
}

@article{ke2025survey,
  title={A survey of frontiers in llm reasoning: Inference scaling, learning to reason, and agentic systems},
  author={Ke, Zixuan and Jiao, Fangkai and Ming, Yifei and Nguyen, Xuan-Phi and Xu, Austin and Long, Do Xuan and Li, Minzhi and Qin, Chengwei and Wang, Peifeng and Savarese, Silvio and others},
  journal={arXiv preprint arXiv:2504.09037},
  year={2025}
}

@article{guo2017critical,
  title={A critical review of cascading failure analysis and modeling of power system},
  author={Guo, Hengdao and Zheng, Ciyan and Iu, Herbert Ho-Ching and Fernando, Tyrone},
  journal={Renewable and Sustainable Energy Reviews},
  volume={80},
  pages={9--22},
  year={2017},
  publisher={Elsevier}
}

@article{xia2025tokenskip,
  title={Tokenskip: Controllable chain-of-thought compression in llms},
  author={Xia, Heming and Leong, Chak Tou and Wang, Wenjie and Li, Yongqi and Li, Wenjie},
  journal={arXiv preprint arXiv:2502.12067},
  year={2025}
}

@article{hou2022token,
  title={Token dropping for efficient bert pretraining},
  author={Hou, Le and Pang, Richard Yuanzhe and Zhou, Tianyi and Wu, Yuexin and Song, Xinying and Song, Xiaodan and Zhou, Denny},
  journal={arXiv preprint arXiv:2203.13240},
  year={2022}
}

@article{pan2024llmlingua,
  title={Llmlingua-2: Data distillation for efficient and faithful task-agnostic prompt compression},
  author={Pan, Zhuoshi and Wu, Qianhui and Jiang, Huiqiang and Xia, Menglin and Luo, Xufang and Zhang, Jue and Lin, Qingwei and R{\"u}hle, Victor and Yang, Yuqing and Lin, Chin-Yew and others},
  journal={arXiv preprint arXiv:2403.12968},
  year={2024}
}

@inproceedings{devlin2019bert,
  title={Bert: Pre-training of deep bidirectional transformers for language understanding},
  author={Devlin, Jacob and Chang, Ming-Wei and Lee, Kenton and Toutanova, Kristina},
  booktitle={Proceedings of the 2019 conference of the North American chapter of the association for computational linguistics: human language technologies, volume 1 (long and short papers)},
  pages={4171--4186},
  year={2019}
}

@inproceedings{ma2025step,
  title={What are step-level reward models rewarding? counterintuitive findings from mcts-boosted mathematical reasoning},
  author={Ma, Yiran and Chen, Zui and Liu, Tianqiao and Tian, Mi and Liu, Zhuo and Liu, Zitao and Luo, Weiqi},
  booktitle={Proceedings of the AAAI Conference on Artificial Intelligence},
  volume={39},
  pages={24812--24820},
  year={2025}
}

@article{hou2504thinkprune,
  title={Thinkprune: Pruning long chain-of-thought of llms via reinforcement learning, 2025},
  author={Hou, Bairu and Zhang, Yang and Ji, Jiabao and Liu, Yujian and Qian, Kaizhi and Andreas, Jacob and Chang, Shiyu},
  journal={URL https://arxiv. org/abs/2504.01296},
  year={2025}
}

@article{tyen2023llms,
  title={LLMs cannot find reasoning errors, but can correct them given the error location},
  author={Tyen, Gladys and Mansoor, Hassan and C{\u{a}}rbune, Victor and Chen, Peter and Mak, Tony},
  journal={arXiv preprint arXiv:2311.08516},
  year={2023}
}

@article{jiang2025cascadia,
  title={Cascadia: A Cascade Serving System for Large Language Models},
  author={Jiang, Youhe and Fu, Fangcheng and Zhao, Wanru and Rabanser, Stephan and Lane, Nicholas D and Yuan, Binhang},
  journal={arXiv preprint arXiv:2506.04203},
  year={2025}
}

@article{team2024qwen2,
  title={Qwen2 technical report},
  author={Team, Qwen},
  journal={arXiv preprint arXiv:2407.10671},
  year={2024}
}

@article{cobbe2021training,
  title={Training verifiers to solve math word problems},
  author={Cobbe, Karl and Kosaraju, Vineet and Bavarian, Mohammad and Chen, Mark and Jun, Heewoo and Kaiser, Lukasz and Plappert, Matthias and Tworek, Jerry and Hilton, Jacob and Nakano, Reiichiro and others},
  journal={arXiv preprint arXiv:2110.14168},
  year={2021}
}

@article{hendrycks2021measuring,
  title={Measuring mathematical problem solving with the math dataset},
  author={Hendrycks, Dan and Burns, Collin and Kadavath, Saurav and Arora, Akul and Basart, Steven and Tang, Eric and Song, Dawn and Steinhardt, Jacob},
  journal={arXiv preprint arXiv:2103.03874},
  year={2021}
}

@inproceedings{lightman2023let,
  title={Let's verify step by step},
  author={Lightman, Hunter and Kosaraju, Vineet and Burda, Yuri and Edwards, Harrison and Baker, Bowen and Lee, Teddy and Leike, Jan and Schulman, John and Sutskever, Ilya and Cobbe, Karl},
  booktitle={The Twelfth International Conference on Learning Representations},
  year={2023}
}

@article{hu2022lora,
  title={Lora: Low-rank adaptation of large language models.},
  author={Hu, Edward J and Shen, Yelong and Wallis, Phillip and Allen-Zhu, Zeyuan and Li, Yuanzhi and Wang, Shean and Wang, Lu and Chen, Weizhu and others},
  journal={ICLR},
  volume={1},
  number={2},
  pages={3},
  year={2022}
}

@article{liu2024can,
  title={Can language models learn to skip steps?},
  author={Liu, Tengxiao and Guo, Qipeng and Hu, Xiangkun and Jiayang, Cheng and Zhang, Yue and Qiu, Xipeng and Zhang, Zheng},
  journal={Advances in Neural Information Processing Systems},
  volume={37},
  pages={45359--45385},
  year={2024}
}

@article{xu2025chain,
  title={Chain of draft: Thinking faster by writing less},
  author={Xu, Silei and Xie, Wenhao and Zhao, Lingxiao and He, Pengcheng},
  journal={arXiv preprint arXiv:2502.18600},
  year={2025}
}

@article{han2024token,
  title={Token-budget-aware llm reasoning},
  author={Han, Tingxu and Wang, Zhenting and Fang, Chunrong and Zhao, Shiyu and Ma, Shiqing and Chen, Zhenyu},
  journal={arXiv preprint arXiv:2412.18547},
  year={2024}
}

@article{ding2025dynamic,
  title={Dynamic parallel tree search for efficient llm reasoning},
  author={Ding, Yifu and Jiang, Wentao and Liu, Shunyu and Jing, Yongcheng and Guo, Jinyang and Wang, Yingjie and Zhang, Jing and Wang, Zengmao and Liu, Ziwei and Du, Bo and others},
  journal={arXiv preprint arXiv:2502.16235},
  year={2025}
}

@article{wadhwa2024investigating,
  title={Investigating mysteries of cot-augmented distillation},
  author={Wadhwa, Somin and Amir, Silvio and Wallace, Byron C},
  journal={arXiv preprint arXiv:2406.14511},
  year={2024}
}

@article{zhu2025breaking,
  title={Breaking the MoE LLM Trilemma: Dynamic Expert Clustering with Structured Compression},
  author={Zhu, Peijun and Yang, Ning and Wei, Jiayu and Wu, Jinghang and Zhang, Haijun},
  journal={arXiv preprint arXiv:2510.02345},
  year={2025}
}

@article{wingate2022prompt,
  title={Prompt compression and contrastive conditioning for controllability and toxicity reduction in language models},
  author={Wingate, David and Shoeybi, Mohammad and Sorensen, Taylor},
  journal={arXiv preprint arXiv:2210.03162},
  year={2022}
}

@misc{xuefei2023skeleton,
  title={Skeleton-ofthought: Prompting LLMs for Efficient Parallel Generation},
  author={Xuefei, Ning and Zinan, Lin and Zixuan, Zhou and others},
  journal={2023-07-28)[2023-11-25]. https://arxiv. org/abs/2307.15337},
  year={2023}
}

@article{kumar2025overthink,
  title={Overthink: Slowdown attacks on reasoning llms},
  author={Kumar, Abhinav and Roh, Jaechul and Naseh, Ali and Karpinska, Marzena and Iyyer, Mohit and Houmansadr, Amir and Bagdasarian, Eugene},
  journal={arXiv preprint arXiv:2502.02542},
  year={2025}
}

@article{liu2025efficient,
  title={Efficient inference for large reasoning models: A survey},
  author={Liu, Yue and Wu, Jiaying and He, Yufei and Gao, Hongcheng and Chen, Hongyu and Bi, Baolong and Gong, Ruihan and Zhang, Jiaheng and Huang, Zhiqi and Hooi, Bryan},
  journal={arXiv preprint arXiv:2503.23077},
  year={2025}
}

@article{wang2024svd,
  title={Svd-llm: Truncation-aware singular value decomposition for large language model compression},
  author={Wang, Xin and Zheng, Yu and Wan, Zhongwei and Zhang, Mi},
  journal={arXiv preprint arXiv:2403.07378},
  year={2024}
}

@article{kamoi2024can,
  title={When can llms actually correct their own mistakes? a critical survey of self-correction of llms},
  author={Kamoi, Ryo and Zhang, Yusen and Zhang, Nan and Han, Jiawei and Zhang, Rui},
  journal={Transactions of the Association for Computational Linguistics},
  volume={12},
  pages={1417--1440},
  year={2024},
  publisher={MIT Press 255 Main Street, 9th Floor, Cambridge, Massachusetts 02142, USA~…}
}

@article{liu2024intrinsic,
  title={Intrinsic self-correction for enhanced morality: An analysis of internal mechanisms and the superficial hypothesis},
  author={Liu, Guangliang and Mao, Haitao and Tang, Jiliang and Johnson, Kristen Marie},
  journal={arXiv preprint arXiv:2407.15286},
  year={2024}
}

@article{kim2025search,
  title={Search-Based Correction of Reasoning Chains for Language Models},
  author={Kim, Minsu and Falet, Jean-Pierre and Richardson, Oliver E and Chen, Xiaoyin and Jain, Moksh and Ahn, Sungjin and Ahn, Sungsoo and Bengio, Yoshua},
  journal={arXiv preprint arXiv:2505.11824},
  year={2025}
}

@article{loshchilov2017decoupled,
  title={Decoupled weight decay regularization},
  author={Loshchilov, Ilya and Hutter, Frank},
  journal={arXiv preprint arXiv:1711.05101},
  year={2017}
}

@article{zheng2024llamafactory,
  title={Llamafactory: Unified efficient fine-tuning of 100+ language models},
  author={Zheng, Yaowei and Zhang, Richong and Zhang, Junhao and Ye, Yanhan and Luo, Zheyan and Feng, Zhangchi and Ma, Yongqiang},
  journal={arXiv preprint arXiv:2403.13372},
  year={2024}
}

@article{dubey2024llama,
  title={The llama 3 herd of models},
  author={Dubey, Abhimanyu and Jauhri, Abhinav and Pandey, Abhinav and Kadian, Abhishek and Al-Dahle, Ahmad and Letman, Aiesha and Mathur, Akhil and Schelten, Alan and Yang, Amy and Fan, Angela and others},
  journal={arXiv e-prints},
  pages={arXiv--2407},
  year={2024}
}

@article{yang2025token,
  title={Token-Importance Guided Direct Preference Optimization},
  author={Yang, Ning and Lin, Hai and Liu, Yibo and Tian, Baoliang and Liu, Guoqing and Zhang, Haijun},
  journal={arXiv preprint arXiv:2505.19653},
  year={2025}
}

@article{zhang2026chain,
  title={Chain-of-Thought Compression Should Not Be Blind: V-Skip for Efficient Multimodal Reasoning via Dual-Path Anchoring},
  author={Zhang, Dongxu and Sun, Yiding and Tan, Cheng and Yan, Wenbiao and Yang, Ning and Zhu, Jihua and Zhang, Hiajun},
  journal={arXiv preprint arXiv:2601.13879},
  year={2026}
}

\appendix

\section{Appendix}
\label{sec:appendix}
This document provides supplementary material, including pseudocode for the ASCoT mechanism, experimental setup, error-injection experiments, and results analysis. It also covers model configurations, hyperparameters used in the study. Additionally, the document discusses the model’s Late-Stage Fragility and outlines current limitations and future directions.

\section{Algorithmic Details}

Algorithm~\ref{alg:ASCoT} formally outlines the inference pipeline of the proposed ASCoT framework. The process initiates with the Intelligent Routing Mechanism (IRM), which semantically prunes the initial chain $\mathcal{C}_{\text{init}}$ to filter redundant tokens, yielding a compressed sequence $\mathcal{C}_{\text{pruned}}$.

Subsequently, the system iterates through each reasoning step $t_k$. At each juncture, the Adaptive Verification Manager (AVM) assesses the step against the accumulated context $S_{\text{ctx}}$, computing a risk score based on both content quality and positional impact. If this score exceeds the predefined threshold $\tau$, the step is flagged as fragile. In such cases, the Multi-Perspective Self-Correction Engine (MSCE) is triggered to generate a rectified step $t_k^{\text{corr}}$ via dual-path correction. Conversely, low-risk steps are retained directly. This selective intervention ensures that computational resources are reallocated specifically to fix late-stage vulnerabilities while maintaining overall efficiency.

\begin{algorithm*}[t]
\caption{ASCoT: Adaptive Self-Correction Chain-of-Thought}
\label{alg:ASCoT}
\begin{algorithmic}[1]
\Require Initial CoT $\mathcal{C}_{\text{init}}$; Compression ratio $\gamma$; Risk threshold $\tau$
\Ensure Corrected CoT $\mathcal{C}_{\text{final}}$

\vspace{0.1cm}
\LINECOMMENT{Stage 1: Semantic Pruning via IRM}
\State $\mathcal{C}_{\text{pruned}} \gets \mathrm{SemanticPruning}(\mathcal{C}_{\text{init}}, \gamma)$

\vspace{0.1cm}
\LINECOMMENT{Stage 2 \& 3: Adaptive Verification and Correction}
\State $\mathcal{C}_{\text{final}} \gets \emptyset$; $S_{\text{ctx}} \gets \emptyset$

\For{\textbf{each} step $t_k$ \textbf{in} $\mathcal{C}_{\text{pruned}}$}
    \LINECOMMENT{Calculate risk via AVM (Logic, Fact, Clarity, Position)}
    \State $R_k \gets \mathrm{ComputeRiskScore}(t_k, S_{\text{ctx}})$

    \If{$R_k > \tau$}
        \LINECOMMENT{High risk detected: Trigger MSCE correction}
        \State $t_k^{\text{corr}} \gets \mathrm{DualPathCorrection}(t_k, S_{\text{ctx}})$
        \State $\mathrm{Append}(t_k^{\text{corr}}, \mathcal{C}_{\text{final}})$ \INLINECOMMENT{Use corrected step}
        \State $S_{\text{ctx}} \gets \mathrm{UpdateContext}(S_{\text{ctx}}, t_k^{\text{corr}})$
    \Else
        \State $\mathrm{Append}(t_k, \mathcal{C}_{\text{final}})$ \INLINECOMMENT{Keep original step}
        \State $S_{\text{ctx}} \gets \mathrm{UpdateContext}(S_{\text{ctx}}, t_k)$
    \EndIf
\EndFor

\vspace{0.1cm}
\LINECOMMENT{Final Output Generation}
\State \textbf{return} $\mathcal{C}_{\text{final}}$

\end{algorithmic}
\end{algorithm*}

\section{Qualitative Analysis of Late-Stage Fragility}
\label{sec:app_fragility}
\begin{figure*}[t]
\centering
\includegraphics[width=1.0\textwidth]{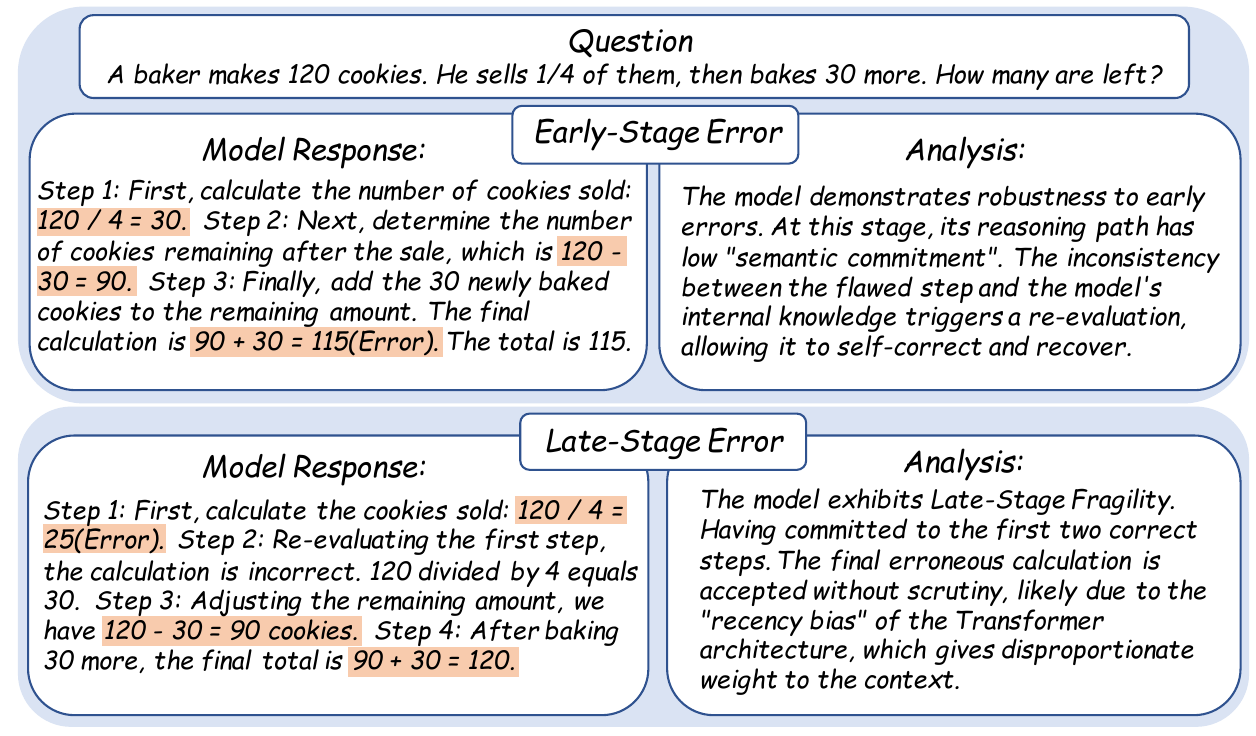} 
\caption{An illustration of Late-Stage Fragility. The model recovers from an early-stage calculation error, but fails when an identical type of error is introduced in the final step. The tokens that occur related errors are highlighted in orange.}
\label{fig4}
\end{figure*}

To visually demonstrate the Late-Stage Fragility phenomenon discussed in Section 4.3, we provide a comparative case study in Figure~\ref{fig4}. The example contrasts the model's robust recovery from early-stage perturbations against its failure to correct identical errors in the final reasoning steps.

\section{Experimental Setup}
\label{appen::Exset}
To clearly elucidate the algorithmic pipeline of the ASCoT methodology, we provide its pseudocode. 
The algorithm commences upon receiving a problem query, which is first processed by the target Large Language Model (LLM) to generate an initial CoT solution, denoted as $CoT_{initial}$. This initial output is subsequently passed to the IRM module for token pruning based on a predefined compression ratio $\gamma$, thereby enhancing efficiency. The compressed CoT then undergoes a step-wise risk assessment by the AVM module. For each step $t_k$, the AVM computes a Risk Score $R(t_k)$, which integrates a multi-dimensional quality assessment $Q(t_k)$ with a Positional Impact score $I(k)$. If this score exceeds a predefined threshold $\tau$, the step is flagged as high risk and is routed to the MSCE module for remediation. The MSCE employs a dual-path strategy, encompassing intrinsic reflection and extrinsic independent generation to correct the identified error, yielding the final, refined $CoT_{final}$.


\textbf{Model Configuration and Hyperparameters}
All experiments were conducted on an Ubuntu 20.04 LTS server equipped with four NVIDIA GeForce RTX 3090 GPUs (24GB each), using PyTorch 1.13.0. We fine‑tuned each base LLM via Low‑Rank Adaptation (LoRA) ~\cite{hu2022lora} with rank $r=8$ and scale factor $\alpha=16$. We utilize LLMLingua-2 ~\cite{pan2024llmlingua} as the token importance metric to generate our compressed CoT training data. IRM is evaluated over discrete compression rates in the range $\{0.5,0.6,0.7,0.8,0.9,1.0\}$. The $I(k) = w_a \cdot e^{\alpha (k - 1)}$ was calibrated by fitting to data from controlled error‑injection trials. All fine‑tuning used best‑practice hyperparameters. We train the models for 5 epochs on both datasets. The  peak learning rate is set to $5e-5$, following a cosine decay schedule. We use AdamW ~\cite{loshchilov2017decoupled} for optimization, with a warmup ratio of 0.1. We implement our training process using the LLaMA-Factory ~\cite{zheng2024llamafactory} library. During inference, the maximum number of tokens set to 512 for GSM8K and 1024 for MATH.






\section{Error‑Injection Experiments}
\label{appen:error}
\subsection{Error Simulation}

To probe the model's resilience in a realistic manner, our error-injection methodology does not introduce arbitrary flaws. Instead, it is designed to simulate the spontaneous reasoning failures we observed in the model's own unguided generations, thereby ensuring the ecological validity of our experiment. We focused on two distinct classes of these naturally occurring failure patterns.

The first, Numeric Errors, simulates arithmetic miscalculations. The injection process for this error type involves identifying a numerical value in a correct reasoning step and slightly altering it, for instance, by adding or subtracting a small, fixed constant value to mimic a common calculation mistake. The second, Symbolic Errors, simulates logical inconsistencies. This is achieved by identifying a primary arithmetic operator within a step and systematically replacing it with its mathematical inverse ($e.g.,$ + is replaced by -, and * by ÷). To maintain procedural integrity, we preemptively implemented constraints to avoid invalid operations, such as division by zero.

Our overall experimental procedure involved systematically injecting one such controlled error at a specific position k within a baseline of validated, correct reasoning chains. Following this injection, the model was prompted to resume the reasoning process from that point, and the accuracy of the final answer was recorded to precisely measure the positional impact of the error.

\begin{table}[t]
\centering
\begin{tabular}{c c c}
\toprule
\textbf{Infer Steps} & \textbf{Counts}& \textbf{Accuracy} \\ 
\midrule
2 & 786 & 98.85 \\ 
3 & 2125 & 96.19 \\ 
4 & 2022 & 95.75\\ 
5 & 1180 & 94.07\\ 
6 & 484 & 90.91\\ 
7 & 137 & 85.40\\
8 & 58 & 67.24\\
9 & 7 & 57.14\\
\bottomrule
\end{tabular}
\caption{Performance analysis by reasoning step count on the GSM8K dataset, illustrating both the frequency of problems per step count and the corresponding model accuracy.}
\label{table4}
\end{table}

\subsection{Distribution of Results}

Table ~\ref{table4} illustrates the proportion of total steps across all problems. We can conclude from the chart that a significant majority of problems are concentrated in the 2-5 step range, with 3-step and 4-step problems being the most prevalent.

When injecting a numerical error, the accuracy drops are relatively uniform across positions: at step 2 yields a drop of 59.92\%, at step 3 a drop of 60.47\%, and at step 4 a drop of 61.70\%a miscalculation. In contrast, symbolic errors induce more severe degradation, and this severity escalates with later injection points. Replacement of an operator in the first error step (2/2) reduces the accuracy by 69.67\%, while an error in the final step (4/4) inflicts a reduction of 80.80\%, with symbolic faults becoming increasingly catastrophic when introduced deeper into the chain.


\textbf{Analysis of Late‑Stage Fragility}\quad
Early-stage errors often activate the model’s latent self-correction capability, wherein subsequent reasoning steps can reinterpret and compensate for previous mistakes, leading to a correct final answer. This is facilitated by the model maintaining higher semantic entropy and contextual flexibility during the initial stages of reasoning. However, as the reasoning chain progresses, the model increasingly commits to a specific semantic trajectory, gradually narrowing its generative space and becoming less sensitive to alternative interpretations. We refer to this phenomenon as semantic commitment, a state in which the model’s outputs become rigidly anchored to previously generated content. At this point, even minor numerical or symbolic errors near the conclusion are less likely to be detected or corrected, resulting in high failure rates. 

Our findings do not aim to refute the cascading failure hypothesis entirely, but rather to refine and contextualize it. Previous work suggests that early-stage errors in CoT reasoning tend to propagate forward, compromising subsequent steps and the final answer. However, our analysis reveals that the impact of reasoning errors is not uniform. We also find that late-stage reasoning steps are disproportionately sensitive to localized perturbations. Errors introduced near the conclusion of reasoning are much less likely to be corrected. In contrast, early-stage errors may still trigger the model’s latent self-correction mechanisms. This indicates that cascading failures do occur, but that Late-Stage Fragility constitutes an additional failure mode not previously emphasized.

\begin{figure*}[t]
\centering
\includegraphics[width=1.0\textwidth]{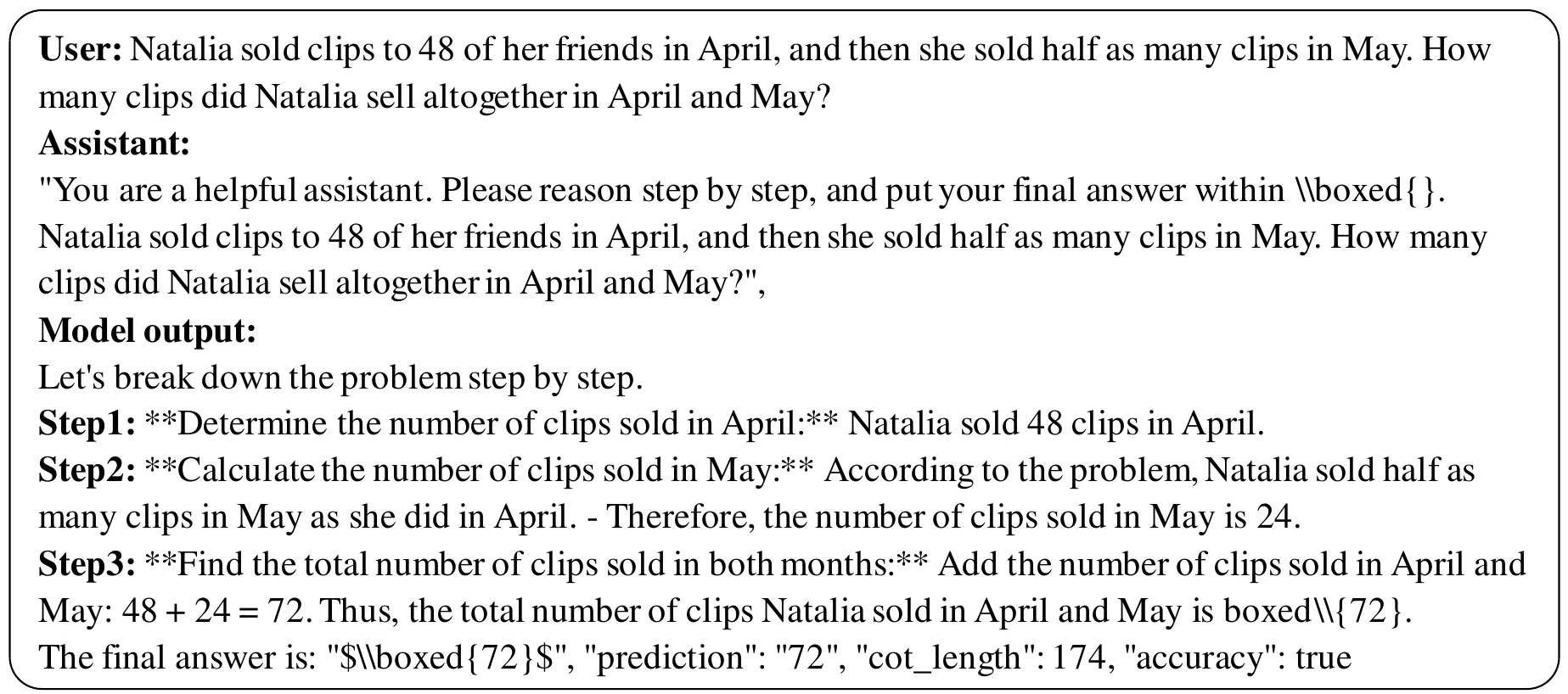} 
\caption{Example generations by ASCoT, using Qwen2.5-Instruct 7B.}
\label{fig7}
\end{figure*}

\section{Detailed Ablation Studies}
\label{sec:app_ablation}

In this section, we present a series of ablation experiments to validate the design choices of ASCoT. All experiments were conducted using the LLaMA-3.1-8B model on the GSM8K benchmark.

\subsection{Impact of Core Components}
We first investigate the necessity of the two fundamental stages: semantic pruning (IRM) and the correction pipeline (AVM+MSCE). Table~\ref{tab:ablation_core} summarizes the results.
\begin{table}[t]
\centering
\caption{Ablation of core components. ``w/o IRM'' denotes removing the pruning stage; w/o Correction denotes using only pruning without any verification or correction.}
\setlength{\tabcolsep}{6pt}
\resizebox{\columnwidth}{!}{%
    \begin{tabular}{l c c c}
    \toprule
    \textbf{Configuration} & \textbf{Acc. (\%)} & \textbf{Tokens} & \textbf{ActRatio} \\
    \midrule
    \textbf{ASCoT (Full)} & \textbf{86.9} & \textbf{214.24} & \textbf{1.00} \\
    w/o IRM & 87.1 & 325.40 & 1.52 \\
    w/o Correction & 82.5 & 198.12 & 0.92 \\
    \bottomrule
    \end{tabular}%
}
\label{tab:ablation_core}
\end{table}
As shown in Table~\ref{tab:ablation_core}, removing the IRM leads to a massive increase in token usage (+52\%) with negligible accuracy gain (+0.2\%), validating IRM's critical role in efficiency. Conversely, removing the correction pipeline (w/o Correction) causes accuracy to drop significantly to 82.5\%, underscoring that semantic pruning alone induces fragility that must be compensated for by our robust verification mechanism.

\subsection{Efficacy of Position-Aware Verification}
A key innovation of ASCoT is the integration of the Positional Impact Score within the AVM module to address Late-Stage Fragility. Table~\ref{tab:ablation_avm} compares our strategy against a uniform risk assessment.
\begin{table}[t]
\centering
\caption{Ablation of the AVM strategy. We compare our position-aware scoring against a uniform weighting scheme.}
\setlength{\tabcolsep}{8pt}
\resizebox{0.9\columnwidth}{!}{%
    \begin{tabular}{l c c}
    \toprule
    \textbf{AVM Strategy} & \textbf{Acc. (\%)} & \textbf{$\Delta$ Acc.} \\
    \midrule
    \textbf{Position-Aware (Ours)} & \textbf{86.9} & -- \\
    Uniform Weights & 84.3 & -2.6 \\
    \bottomrule
    \end{tabular}%
}
\label{tab:ablation_avm}
\end{table}
The results in Table~\ref{tab:ablation_avm} reveal a 2.6\% performance degradation when position awareness is removed. This empirical evidence supports our hypothesis that errors in later reasoning steps are more detrimental. By assigning higher risk weights to these terminal steps, ASCoT effectively reallocates verification resources to where they are most needed.

\subsection{Analysis of Dual-Path Correction}
Finally, we dissect the Multi-Perspective Self-Correction Engine (MSCE) to evaluate the contribution of its Dual-Path (Intrinsic + Extrinsic) strategy compared to single-source baselines.
\begin{table}[t]
\centering
\caption{Ablation of the MSCE correction strategy. ``Intrinsic'' relies on self-reflection; ``Extrinsic'' uses independent regeneration.}
\setlength{\tabcolsep}{8pt}
\resizebox{0.95\columnwidth}{!}{%
    \begin{tabular}{l c c}
    \toprule
    \textbf{Correction Strategy} & \textbf{Acc. (\%)} & \textbf{$\Delta$ Acc.} \\
    \midrule
    \textbf{Dual-Path (Ours)} & \textbf{86.9} & -- \\
    Extrinsic Only & 86.1 & -0.8 \\
    Intrinsic Only & 85.0 & -1.9 \\
    \bottomrule
    \end{tabular}%
}
\label{tab:ablation_msce}
\end{table}
Table~\ref{tab:ablation_msce} demonstrates that the Dual-Path strategy outperforms individual approaches. Relying solely on Intrinsic correction yields the lowest accuracy (85.0\%), likely because the model struggles to escape its own erroneous reasoning context. Extrinsic correction performs better (86.1\%) but is surpassed by the synergistic combination of both, which leverages the strengths of context-aware reflection and objective regeneration.

\section{Generation Examples}
To provide a qualitative illustration of our method's end-to-end functionality, we present representative examples of reasoning chains after processing by the ASCoT pipeline. Figure~\ref{fig7} displays the corrected output after being processed by ASCoT. This transformation from a flawed to a correct reasoning chain highlights the core capabilities of our system. It demonstrates the AVM module successfully identifying a high risk step by analyzing its content quality and positional impact, followed by the MSCE module executing a targeted repair. This example confirms ASCoT's ability to structure reasoning but, more importantly, to dynamically detect and rectify internal inconsistencies, ultimately leading to a more trustworthy final answer.

\section{Future Work}

We propose several promising future directions. A key avenue is the augmentation of MSCE module through integration with external, deterministic oracles like calculators, code interpreters, or knowledge-base APIs. By querying these verifiable tools, the correction engine could access ground-truth feedback, significantly enhancing its reliability and moving beyond the constraints of the model’s internal knowledge. Another important direction involves developing online adaptive learning mechanisms. Such a system would enable ASCoT to continuously self-supervise and refine its performance in real time across various domains, evolving from a static framework into a more dynamic and robust reasoning system.

\end{document}